\documentclass[letterpaper, 10 pt, journal, twoside]{IEEEtran}
\usepackage{graphicx}
\usepackage[caption=false,font=footnotesize]{subfig}  
\usepackage{cite}
\usepackage{booktabs} 
\usepackage{multirow}
\usepackage{threeparttable}  
\usepackage{amssymb}  
\usepackage{amsmath}
\usepackage{gensymb}  
\usepackage{colortbl}  

\makeatletter
\let\NAT@parse\undefined
\makeatother
\usepackage[colorlinks,
            allcolors=blue,
            linkcolor=blue,
            anchorcolor=blue,
            citecolor=blue]{hyperref}  

\ifCLASSINFOpdf
\else
\fi
\newcommand{\secref}[1]{Section~\ref{#1}}
\newcommand{\figref}[1]{Fig.~\ref{#1}}
\newcommand{\tabref}[1]{Table~\ref{#1}}

\begin{document}
%
\title{GOPT: Generalizable Online 3D Bin Packing via Transformer-based Deep Reinforcement Learning}
%
%
%

\author{Heng Xiong$^{1}$, Changrong Guo$^{1}$, Jian Peng$^{1}$, Kai Ding$^{2}$, Wenjie Chen$^{3,4}$, Xuchong Qiu$^{2}$, Long Bai$^{1}$, and Jianfeng Xu$^{1,5}$%
\thanks{Manuscript received: June 22, 2024; Accepted September 7, 2024.}
\thanks{This paper was recommended for publication by Editor Chao-Bo Yan upon evaluation of the Associate Editor and Reviewers' comments.
This work was supported by the National Key R\&D Program of China under Grant No. 2022YFB4700300. \textit{(Corresponding author: Jianfeng Xu.)}} 
\thanks{$^{1}$Heng Xiong, Changrong Guo, Jian Peng, and Long Bai are with State Key Laboratory of Intelligent Manufacturing Equipment and Technology, School of Mechanical Science and Engineering, Huazhong University of Science and Technology, Wuhan 430074, China
        {\tt\footnotesize \{xiongheng, guochangrong, peng\_jian, bailong\}@hust.edu.cn}}%
\thanks{$^{2}$Kai Ding and Xuchong Qiu are with BOSCH Corporate Research, China
        {\tt\footnotesize \{firstname.lastname\}@cn.bosch.com}}%
\thanks{$^{3,4}$Wenjie Chen is with State Key Laboratory of High-end Heavy-load Robots, Midea Group, Foshan 528300, China, and also with Midea Corporate Research Center, Foshan 528311, China
        {\tt\footnotesize chenwj42@midea.com}}%
\thanks{$^{5}$Jianfeng Xu is with State Key Laboratory of Intelligent Manufacturing Equipment and Technology, School of Mechanical Science and Engineering, Huazhong University of Science and Technology, Wuhan 430074, China, and also with HUST-Wuxi Research Institute, Wuxi 214174, China
        {\tt\footnotesize jfxu@hust.edu.cn}}%
\thanks{Digital Object Identifier (DOI): see top of this page.}
}

%
%

\markboth{IEEE Robotics and Automation Letters. Preprint Version. Accepted September, 2024}
{Xiong \MakeLowercase{\textit{et al.}}: GOPT: Generalizable Online 3D Bin Packing via Transformer-based Deep Reinforcement Learning}

%



\maketitle

\begin{abstract}

Robotic object packing has broad practical applications in the logistics and automation industry, often formulated by researchers as the online 3D Bin Packing Problem (3D-BPP). However, existing DRL-based methods primarily focus on enhancing performance in limited packing environments while neglecting the ability to generalize across multiple environments characterized by different bin dimensions. To this end, we propose GOPT, a \underline{g}eneralizable \underline{o}nline 3D Bin \underline{P}acking approach via \underline{T}ransformer-based deep reinforcement learning (DRL). First, we design a Placement Generator module to yield finite sub-spaces as placement candidates and the representation of the bin. Second, we propose a Packing Transformer, which fuses the features of the items and bin, to identify the spatial correlation between the item to be packed and available sub-spaces within the bin. Coupling these two components enables GOPT's ability to perform inference on bins of varying dimensions. We conduct extensive experiments and demonstrate that GOPT not only achieves superior performance against the baselines, but also exhibits excellent generalization capabilities. Furthermore, the deployment with a robot showcases the practical applicability of our method in the real world. The source code will be publicly available at \url{https://github.com/Xiong5Heng/GOPT}.

\end{abstract}


%
\IEEEpeerreviewmaketitle



\section{Introduction}

\IEEEPARstart{W}{ith} the prosperity of the global trade and e-commerce market, warehouse automation has developed rapidly in recent years. Efficient object placement in the warehouse through optimal packing strategies can bring numerous benefits, such as reduced labor requirements and cost savings~\cite{wang2021dense}. 

\begin{figure} [!t]
    \centering
    \includegraphics[width=0.8\linewidth]{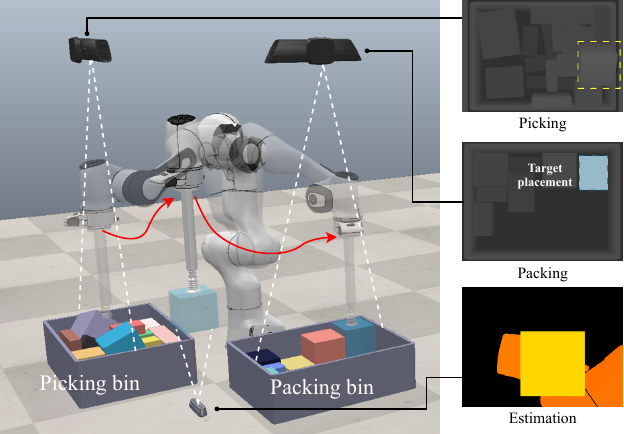} 
    \caption{Robot picking and packing pipeline. 
        Left: A robot randomly picks an item from a cluttered collection of boxes and packs it in a compact manner, and three RGB-D cameras are mounted.
        Right: Two overhead cameras observe the status of the two bins, respectively, and one up-looking camera estimates the dimension of the picked item.
        } 
    \label{fig:pipeline}
    \vspace{-0.3cm}
\end{figure}

\figref{fig:pipeline} illustrates an example of item picking and packing using a robotic arm. In this paper, it is assumed that the robot picking is well implemented. Researchers have commonly addressed the placement challenge in robot packing by formulating it as an online 3D Bin Packing Problem (3D-BPP) \cite{zhao2021online, xiong2023towards}. As one of the classic combinatorial optimization problems, 3D-BPP strives to place a set of known cuboid items in an axis-aligned fashion into a bin to maximize space utilization. However, observing all items and obtaining full knowledge about them is challenging in many real-world scenarios. The online 3D-BPP is a more practical variant of 3D-BPP that refers to packing items one by one under the observation of only the incoming item. 

Due to the limited knowledge, the online 3D-BPP cannot be solved by exact algorithms \cite{do2021practical}. Researchers have previously focused on developing heuristics with the greedy objectives for the problem, which are designed by abstracting the experience of human packers \cite{ha2017online}. However, while intuitive, these heuristics typically yield sub-optimal solutions. In recent years, there has been an emerging research interest in resolving online 3D-BPP via deep reinforcement learning (DRL) \cite{zhao2021online, verma2020generalized, yang2021packerbot, xiong2023towards}, and indeed, DRL-based methods demonstrate impressive performance. Nevertheless, it is noteworthy that the training process often encounters challenges in reaching convergence \cite{zhao2021online, zhao2021learning}, and these methods struggle to generalize effectively across diverse packing scenarios, especially those characterized by different bin dimensions. These limitations substantially curtail the broader applicability of DRL in typical use cases. More specifically, the current state-of-the-art DRL-based methods can only perform inference on bins of the same size as those they are trained on \cite{xiong2023towards, yang2023heuristics}. Trained models are not transferable to bins of different sizes. Additionally, the inherent dependence of the packing action space size on bin dimensions in these methods presents significant challenges for model convergence, especially when dealing with larger bins \cite{zhao2022learning}.

Motivated by the aforementioned limitations, this paper proposes GOPT, a generalizable online 3D Bin Packing approach via Transformer-based DRL, as shown in \figref{fig:overview}. In GOPT, a \textit{Placement Generator (PG)} module first adopts a heuristic to generate a fixed-length set of free sub-spaces within the current bin as placement candidates, which ensures controllability over the size of the packing action space. Both the placement candidates and the item to be packed are collectively defined as the state of the Markov Decision Process (MDP). Then, GOPT incorporates a novel packing policy network that integrates a \textit{Packing Transformer} module. This module enhances GOPT's generalizability by intrinsically identifying the spatial correlation between the current item and the available sub-spaces, as well as the relations among these sub-spaces, which are derived from the PG module. The Packing Transformer employs self-attention layers and bi-directional cross-attention layers to extract features as inputs to the reinforcement learning policy.

Experiments show that our method outperforms the state-of-the-art packing methods in terms of space utilization and the number of packed objects. 
To the best of our knowledge, our work is the first to provide the generalization capability to infer across various bins with a trained model while maintaining high performance. We also deploy our packing planning method in a robotic manipulator to demonstrate its practical applicability in the real world. 

In summary, our main contributions are: 
(1) GOPT, a novel method for online 3D-BPP that enlarges the packing performance and generalization; (2) A Placement Generator module to modulate the packing action space and represent the state of the bin; (3) A network called Packing Transformer, which captures the relations between the current item and the available sub-spaces, as well as interrelations among sub-spaces; (4) Extensive experimental evaluations comparing GOPT with baselines.

\section{Related Work}

The 3D-BPP is a classical optimization problem and is known to be strongly NP-hard \cite{ali2022line}. 
We herein briefly review related heuristic and DRL-based methods.

\subsection{Heuristic Methods}

Early works primarily focus on designing efficient heuristics for their simplicity. Researchers attempt to define some packing rules distilled from human workers' experience, such as First Fit~\cite{dosa2013first}, Best Fit~\cite{dosa2014optimal}, and Deepest-Bottom-Left-Fill~\cite{wang2010two}. Corner points (CP)~\cite{martello2000three}, extreme points (EP)~\cite{crainic2008extreme}, empty maximal spaces (EMS)~\cite{parreno2008maximal}, and internal corners point (ICP)~\cite{agarwal2020jampacker} endeavor to represent potential free spaces where items can be packed for enhancing heuristic methods. For instance, Ha et al.~\cite{ha2017online} propose OnlineBPH, which selects one EMS to minimize the margin between the faces of the item to be packed and the faces of the EMS. Yarimcam et al.~\cite{yarimcam2014heuristic} provide heuristics expressed in terms of policy matrices by employing the Irace parameter tuning algorithm~\cite{lopez2016irace}. 
Wang et al. \cite{wang2019stable} propose Heightmap-Minimization (HM) which favors the placement that minimizes occupied volume.
To mitigate the uncertainties originating from the real world, Shuai et al. keep deformed boxes stacked close to enhance stability~\cite{shuai2023compliant}. Hu et al. develop a Maximize-Accessible-Convex-Space (MACS) strategy to optimize the available empty space for packing potential large future items \cite{hu2020tap}. These methods are intuitive and effective; however, they rely on hand-crafted rules and lack the capacity to demonstrate superior performance consistently across diverse problem settings. 
Our work draws on the representation of empty spaces in heuristics, but uses DRL to learn packing patterns without being limited by domain expert knowledge.
\vspace{-0.2cm}

\subsection{DRL-based Methods}

DRL has shown promise in solving certain combinatorial optimization problems~\cite{kool2018attention, nazari2018reinforcement}. Therefore, there is a trend to use DRL to solve the 3D-BPP recently. 
Que et al.~\cite{que2023solving} tackle the offline 3D-BPP with variable height by using DRL with Transformer structure to sequentially address subtasks of position, item selection, and orientation. Instead, we focus on the online 3D-BPP and determine the position and orientation simultaneously. 
To the best of our knowledge, Deep-Pack~\cite{kundu2019deep} is the first to use a DRL-based model to solve a 2D online packing problem, with potential extensions to the online 3D-BPP. It takes an image showing the current state of the bin as input and outputs the pixel location for packing the incoming item. Verma et al.~\cite{verma2020generalized} combine a search heuristic with DRL and propose a two-step strategy for solving the problem with any number and size of bins. 
Zhao et al.~\cite{zhao2021online, zhao2022learning} formulate the problem as a constrained MDP and adopt ACKTR method \cite{wu2017scalable} to train a CNN-based DRL agent. In \cite{zhao2021online}, the DRL agent comprises an actor, a critic, and a predictor to estimate action probabilities, value, and feasibility mask, respectively. It is then improved by decomposing the packing action into the length and width dimensions and orientation to reduce action space~\cite{zhao2022learning}.
They subsequently introduce the Packing Configuration Tree (PCT) based on heuristic search rules and incorporate it into a DRL agent~\cite{zhao2021learning}. The agent employs Graph Attention Networks \cite{velivckovic2017graph} as the policy and is also trained with ACKTR. To investigate the synergies of heuristics and DRL, Yang et al.~\cite{yang2021packerbot} propose PackerBot, which utilizes heuristic reward to assist the DRL agent to perform better. Xiong et al.~\cite{xiong2023towards} introduce a candidate map mechanism to reduce the complexity of exploration and improve performance for the CNN-based DRL agent trained with A2C \cite{mnih2016asynchronous}.
These methods usually concatenate features of the item and the bin directly to learn policies. In contrast, GOPT first proposes free sub-spaces within a bin and utilizes a modified Transformer to discern the relations among these spaces and the relations between them and the current item. Our method ensures generalizability across diverse packing environments.

\section{Methodology}

\begin{figure*}[!t]
    \centering
    \subfloat[The framework of GOPT]{
        \includegraphics[width=0.72\textwidth]{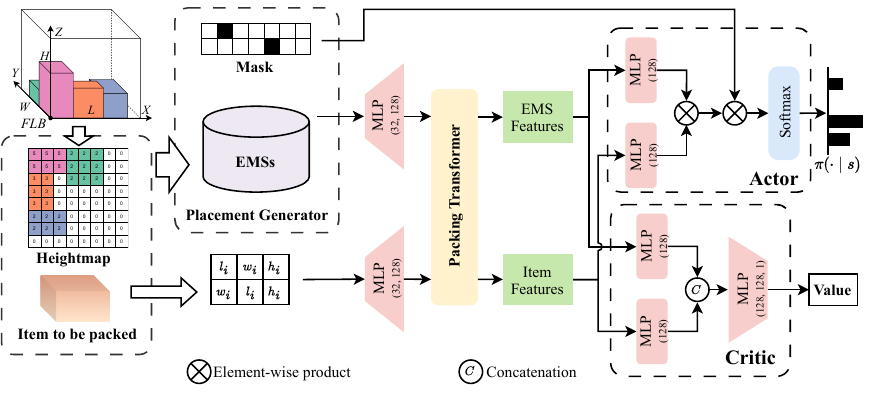}
        \label{fig:overview_a}
    }
    \hfill
    \subfloat[Packing Transformer]{
        \includegraphics[width=0.25\textwidth]{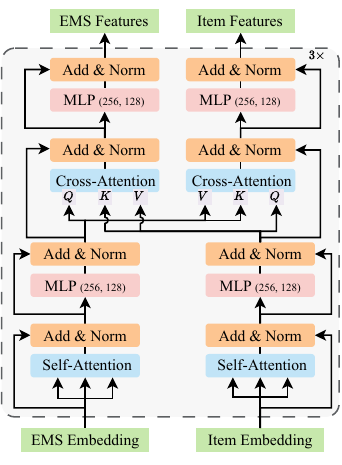}
        \label{fig:overview_b}
    }
    \hfill
    \caption{Overview of our method. 
    (a) In the GOPT, the inputs comprise the item to be packed and the current heightmap of the bin, wherein each cell's value represents the respective height. Utilizing the Placement Generator, a set of EMSs is produced, along with a pairwise action mask between each EMS and the optional orientation of the item. After that, we separately encode the EMSs and the item and then fuse the features using the Packing Transformer, of which outputs are fed into the actor and critic networks to generate logits of all actions and estimate the state-value function; 
    (b) depicts the details of the proposed Packing Transformer. The transformer comprises three stacked blocks, each containing two self-attention and two cross-attention layers.}  
    \label{fig:overview}
\end{figure*}

\subsection{Problem Description}
\label{sec:problem}

As shown in \figref{fig:pipeline}, a robot randomly picks an object from an unstructured pile with a set of box-shaped items of various dimensions. The complete knowledge about all items is unavailable in advance. One camera measures the dimensions of the picked item, which is then placed into the packing bin. This specific scenario can be characterized as an online 3D-BPP. The objective is to place as many items into the bin as possible and maximize the bin's space utilization. 

We define the front-left-bottom (FLB) vertex of the bin with dimensions $(L, W, H)$ as the origin $(0, 0, 0)$, and the directions along the length, width, and height as $X$, $Y$, and $Z$ directions, respectively, as shown in \figref{fig:overview_a}.
For items, $(x_t, y_t, z_t)$ denotes the FLB coordinate of the $t$-th item with dimensions $(l_t, w_t, h_t)$.

In the robot packing task, the following physical constraints must be taken into consideration.

\textit{Orthogonal placement:} Items are placed orthogonally into the bin, and their sides are aligned with the bin's sides.

\textit{Optional orientation:} Items are placed in an upright manner; in conjunction with the first constraint, items have just two distinct vertical in-plane orientations, either $0\degree$ or $90\degree$.

\textit{Static stability:} During the process of packing, items must remain stable under gravity and inter-item forces. For computational efficiency, an item is considered stable if the projection of its geometric center onto its bottom falls inside the support polygon which is formed by the convex hull of all horizontal support points of this item \cite{hu2020tap}.

\subsection{Placement Generator} 
\label{sec:pg}

For the selected item to be packed, we predict the horizontal position $(x_t, y_t)$ and the corresponding orientation of its placement in the bin. The vertical position $z_t$ is analytically determined by the lowest placement position due to gravity. As aforementioned, there are two possible orientations for one item. Therefore, when placing an item into a bin with dimensions $(L, W, H)$, it results in a total number of $L \times W \times 2$ possible placements \cite{zhao2021online}. On the one hand, this quantity is unbearable for the packing problem with the sequential-decision nature because it will grow exponentially with larger bin dimensions. On the other hand, some are inevitably unproductive for the item to be packed within this placement set.

\begin{figure} [!t]  %
    \centering
    \includegraphics[width=0.95\linewidth]{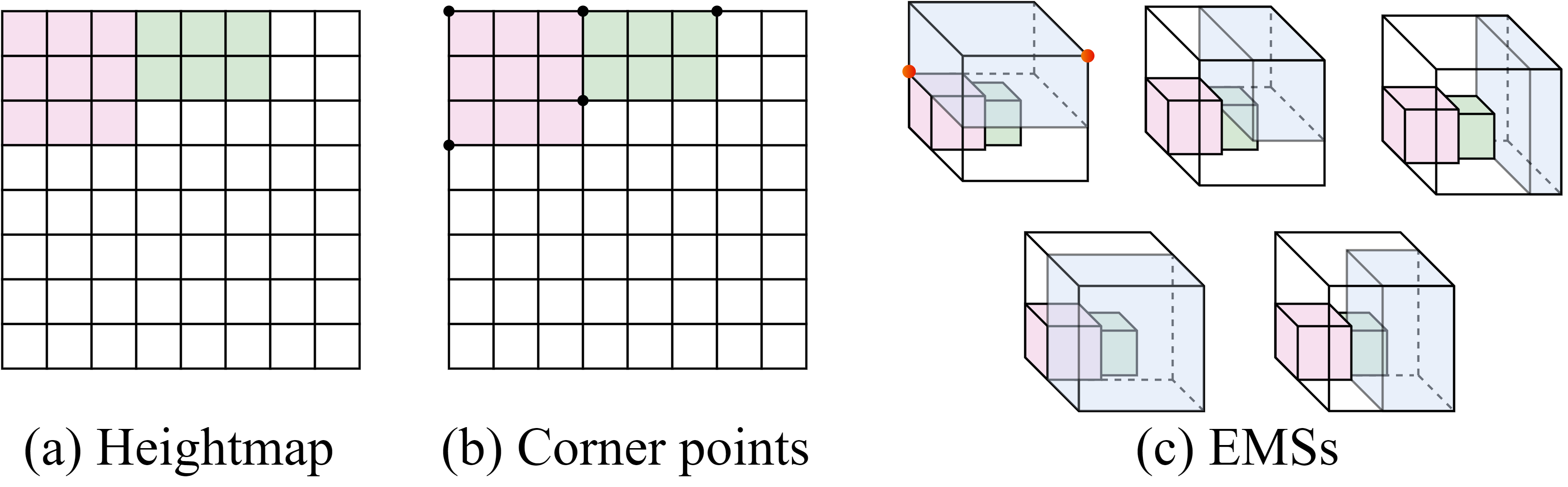}
    \caption{Illustration of the EMS generation procedure. 
    (a) In an example scene with two placed items, the heightmap indicates the current height of stacked items in each grid cell; 
    (b) Five corner points (black dots) are detected at this heightmap; 
    (c) Based on these points, the corresponding largest inscribed rectangles (blue) within the bin are generated, namely EMSs. Taking the first EMS as an example, it is defined by two red vertices of the blue rectangles. }
    \label{fig:ems}
\end{figure}

With the aim of constraining the potentially large placement search space, we design a Placement Generator (PG) module to produce a finite and efficient placement subset based on the incoming item and current bin configuration. We first explicitly represent the real-time status of the bin by utilizing the heightmap. Other methods that leverage planned placements for previous items as the representation \cite{zhao2021learning} lack feedback and closed-loop control. In contrast, the heightmap can be derived from the visual observation captured by a camera conveniently when deploying PG in a real-world robot packing task. Drawing from the empty maximal space (EMS) scheme for managing the empty spaces in a bin \cite{parreno2008maximal, xu2023neural}, candidate placements are computed based on the current state. Specifically, we identify corner points by detecting height variation along the heightmap's $X$ and $Y$ directions. EMSs are then generated by extending unit rectangles from each corner and halting when encountering higher elevation (\figref{fig:ems}). Each EMS can be defined by its FLB vertex and the corresponding opposite vertex as depicted in \figref{fig:ems}c. The resulting 6-dimensional vector is normalized to $[0, 1]$, regardless of the dimensions of the bin. We obtain an EMS subset with controllable size and rank them by height value, denoted as $\{E_i\}_{i=1}^N$. Finally, given an item to be packed, we check the feasibility of each EMS following \secref{sec:problem} and produce a pairwise mask between each EMS and orientation. When packing the item within the bin, we select an appropriate EMS and orientation and align the item's and the EMS's FLB vertices.

\subsection{Reinforcement Learning Formulation}

DRL problems are commonly modeled as a Markov Decision Process (MDP). An MDP with parameters $\left \langle \mathcal{S}, \mathcal{A}, P, R, \gamma \right \rangle$ is utilized to characterize the packing environment in this paper, where $\mathcal{S}$ denotes the state space, $\mathcal{A}$ denotes the action space, $P: \mathcal{S} \times \mathcal{A} \times \mathcal{S} \rightarrow [0, +\infty )$ stands for the transition probabilities, $R: \mathcal{S} \times \mathcal{A} \rightarrow \mathbb{R}$ is the scalar reward function, and $\gamma \in (0, 1]$ is the discount factor for balancing the near-term and long-term rewards in DRL. Reinforcement learning algorithms aim to learn a policy $\pi: \mathcal{S} \times \mathcal{A} \rightarrow \mathbb{R}$, which determines the probability of selecting an action $a$ given a state $s$. The objective of the policy is to maximize the cumulative discounted reward over an episode, expressed as ${\textstyle \sum_{t}^{} \gamma^t r_{t}}$, where $t$ denotes the time step, and $r_t$, $a_t$, and $s_t$ represent the reward, action, and state at time step $t$, respectively. 
In the following, we formulate the online 3D-BPP as an MDP for DRL training.

\textbf{State:} 
At each time step $t$, the policy receives a state $s_t$, comprising the incoming item to be packed $s_{t, item}$ and the current bin configuration $s_{t, bin}$. For the first part, the dimension of the item $(l_t, w_t, h_t)$ is essential. Some studies \cite{xiong2023towards, yang2021packerbot} employ this three-dimensional vector explicitly as the item representation, while others prefer a three-channel map for the convenience of neural network design \cite{zhao2021online, yang2023heuristics}. In the map representation, each channel is assigned $l_t$, $w_t$, and $h_t$, respectively. To account for both the geometry and optional orientations, we propose an item representation which is a $2 \times 3$ matrix,
$s_{t, item} = \begin{bmatrix}
  l_t &w_t &h_t\\
  w_t &l_t &h_t
\end{bmatrix}$, 
where $(l_t, w_t, h_t)$ and $(w_t, l_t, h_t)$ represent the dimensions of the item after rotating it by $0\degree$ and $90\degree$. For the second part, the existing methods include the heightmap \cite{xiong2023towards}, the list of packed items \cite{zhao2021learning}, and the weighted 3D voxel grid \cite{yang2023heuristics}. We choose to leverage the proposed PG (\secref{sec:pg}) to produce a sequence of EMSs satisfying placement constraints as the bin's configuration. The sequence is padded or clipped to a fixed length $N$ with dummy EMSs, i.e. $s_{t, bin} = \{E_i\}_{i=1}^N$. 

\textbf{Action:}
Given the packing state $s_t = (s_{t, item}, s_{t, bin})$, the action $a_t$ involves selecting both an orientation and an EMS for the current item from the sequence of available EMSs. The size of the action space $\mathcal{A}$ depends solely on the length of the sequence and the number of optional orientations, i.e., $\| \mathcal{A} \| = 2N$, irrespective of the bin dimensions. During training, we select the action $a_t$ according to the probability distribution over actions $\pi(\cdot \mid s_t)$, where $\cdot$ represents the set of all possible placements in $s_t$. During testing, we select the action in a deterministic manner by choosing the placement with maximum probability in $\pi(\cdot \mid s_t)$. Note that the probability distribution applying the pairwise action mask between EMSs and orientations ensures that the policy samples valid actions unless no EMS satisfies the constraints.

\textbf{State-Transition:}
In our setting, the transition model is assumed to be deterministic, implying that a specific pair $(s_t, a_t)$ consistently leads to the same subsequent state $s_{t+1}$.

\textbf{Reward:}
The target of the packing problem is to maximize the space ratio of the bin. Hence, we formulate the reward as the step-wise enhancement in space utilization, represented as $r_t = \frac{l_t \cdot w_t \cdot h_t}{L \cdot W \cdot H}$. This dense reward encourages the DRL agent to perform more steps in an episode, thereby leading to more packed items and greater space utilization.   %

\subsection{Network Architecture}

The design of a neural network architecture for the DRL agent is important because the chosen architecture affects the agent's learning and generalization capabilities across varied environments. A simplistic network would be to concatenate the bin and item representations \cite{zhao2021online} or embeddings \cite{yang2021packerbot}. However, this method results in a model whose convolutional and linear layer sizes are contingent upon the dimensions of the bin, rendering the trained model impractical for application across different bins. 

To overcome the challenge of generalization, we propose an attention-based network architecture that focuses on the correlation between the item and the bin's partial spaces. As illustrated in \figref{fig:overview_a}, this architecture comprises three primary components: the Packing Transformer, the actor network, and the critic network. Our network takes the bin representation $s_{t, bin} \in \mathbb{R}^{N \times 6}$ (i.e., a sequence of EMSs from PG) and the item representation $s_{t, item} \in \mathbb{R}^{2 \times 3}$ (i.e., item's dimensions) as inputs. These inputs are then individually processed by Multi-Layer Perceptrons (MLP), which are two-layer linear networks with LeakyReLU activation function. The embedding dimensions of both EMS and the item are set to 128. Subsequently, we then extract features from the embeddings using the designed Packing Transformer, inspired by cross-modality learning across language and vision \cite{li2021selfdoc}. The EMS and item features are then fed into the actor network to generate a probability distribution of potential actions, and fed into the critic network to estimate the expected cumulative reward based on the current state.  

\textbf{Packing Transformer} is depicted in detail in \figref{fig:overview_b}. It is constructed by stacking multiple (three in practice) identical encoder blocks, each containing two self-attention layers, one bi-directional cross-attention layer, and four MLP blocks of two layers comprising $\{128, 128\}$ neurons. The bi-directional cross-attention layer consists of two unidirectional cross-attention layers, one from EMS to item and the other from item to EMS. Residual connections and layer normalization (Norm) are applied after each layer. The self-attention layers play an important role in establishing the intrinsic connections between EMSs or item dimensions, while the bi-directional cross-attention layer facilitates the discovery of inner-relationships from one to another.

\textbf{Actor and critic networks} are both implemented with the MLP layers shown in \figref{fig:overview_a}. In the actor network, both the EMS and item features are processed through an MLP, and the results are multiplied to compute a score map of actions. This is followed by an element-wise multiplication with the action mask to eliminate infeasible actions.

\subsection{Training Method}

We employ the Proximal Policy Optimization (PPO) algorithm \cite{schulman2017proximal} to train the proposed GOPT. PPO is a popular on-policy reinforcement learning algorithm that alternates between collecting data via interactions with the environment and optimizing the following objective, which is approximately maximized in each iteration: 
\begin{equation}
    \mathcal{L}(\theta)=\hat{\mathbb{E}}_{t}[\mathcal{L}^{CLIP}(\theta)-c_{1} \mathcal{L}^{VF}(\theta) + c_{2} S(\pi_{\theta}(\cdot \mid s_{t}))]
\end{equation}
where $\theta$ represents the network parameters, $c_1$, $c_2$ are coefficients, $\mathcal{L}^{CLIP}(\theta)$ is the clipped surrogate objective, $\mathcal{L}^{VF}(\theta)$ is the squared-error loss for the value function, and $S$ denotes the entropy of the policy. Specifically, the surrogate objective is defined as: 
\begin{equation}
    \mathcal{L}^{CLIP} = \hat{\mathbb{E}}_t[\min(p_{t}(\theta)\hat{A}_t, \text{clip}(p_{t}(\theta), 1 - \epsilon, 1 + \epsilon)\hat{A}_t)]
\end{equation}
where $p_{t}(\theta) = \frac{\pi_{\theta}\left(a_{t} \mid s_{t}\right)}{\pi_{\theta_{\text{old}}}\left(a_{t} \mid s_{t}\right)}$ is the action probability ratio between the current policy and the old policy, $\hat{A}_t$ is the estimation of the advantage function which we use Generalized Advantage Estimator (GAE) \cite{schulman2015high} method to compute, and $\epsilon$ indicates the clipped ratio which is used to limit the volume of update and stabilize learning procedure.

\section{Experiments}

\subsection{Implementation Details}
Our method is implemented utilizing PyTorch and adopts the PPO algorithm within the Tianshou framework \cite{weng2022tianshou} for policy training. The maximum number of EMS is set to 80 during each packing step. We train the policy for 1000 epochs and collect a total of 40,000 environment steps over 128 parallel environments in every epoch. Policy updates occur after every 640 environment steps (calculated as $5 \times 128$ steps), with a batch size of 128. 
The Adam optimizer, coupled with a linearly descending learning rate scheduler starting from $7\times 10^{-5}$ is utilized for optimization. 
In terms of PPO loss calculation, the coefficients for value and entropy loss $c_1$, $c_2$ are 0.5 and 0.001, respectively, and the clipped ratio $\epsilon$ is 0.3. The discount factor $\gamma$ is set to 1 to consider future and immediate rewards equally important. For policy updates, we use GAE with $\lambda_{GAE} =0.96$. Our policy training is conducted on a computer equipped with an NVIDIA GeForce RTX 3090 and an Intel Core i7-14700K CPU, reaching convergence from scratch in about six hours.

For experimental validation, we utilize the RS dataset \cite{zhao2021online} for training and evaluating our DRL agent. The bin dimensions $L\times W \times H$ are set to $10\times10\times10$, and the dimensions of items follow $\frac{min(L, W, H)}{10} \le l_t, w_t, h_t \le \frac{min(L, W, H)}{2}$. The dataset comprises 125 types of heterogeneous items, and sequences are dynamically generated by bootstrap sampling during training to reflect the variability in practical scenarios. An additional set of 1000 instances is generated for evaluation purposes, and the average performance is recorded.

\subsection{Performance Evaluation}

\subsubsection{Baselines}

To illustrate the superiority of our method, we select representative methods with publicly available implementations as baselines. We categorize these methods into two groups. The first group consists of four heuristic methods: OnlineBPH~\cite{ha2017online}, Best Fit based on EP~\cite{crainic2008extreme} that packs item in the lowest extreme point, MACS~\cite{hu2020tap}, and HM~\cite{wang2019stable}. The second comprises three DRL-based methods: Zhao et al.~\cite{zhao2021online}, PCT \cite{zhao2021learning}, and Xiong et al. \cite{xiong2023towards}. 
All methods are implemented and executed on the same desktop computer to ensure fair and rigorous comparisons. Furthermore, the DRL-based methods are trained from scratch with an equivalent number of steps, specifically 40 million, to eliminate training disparity bias. 

\begin{table}[!t]
    \centering
    \begin{threeparttable}
    \caption{Performance comparison on a $10\times10\times10$ bin along with the results of the ablation studies.}
    \label{tab:result}
    \setlength{\tabcolsep}{5mm}
    \begin{tabular}{@{\hspace{4mm}}lccc@{\hspace{4mm}}}
        \toprule
        Method                              & \textit{Uti}      & \textit{Sta}    & \textit{Num}   \\ \midrule
        \textit{\textbf{Heuristic}} \\
        OnlineBPH~\cite{ha2017online}       & 51.6\%            & 0.142           & 20.5   \\
        Best Fit~\cite{crainic2008extreme}  & 57.9\%            & 0.124           & 22.9   \\
        MACS~\cite{hu2020tap}               & 50.6\%            & 0.171           & 19.6   \\
        HM~\cite{wang2019stable}            & 56.5\%            & 0.105           & 22.1   \\ \midrule
        \textit{\textbf{DRL-based}}\\
        Zhao et al.~\cite{zhao2021online}   & 70.9\%            & 0.079           & 27.5   \\
        PCT~\cite{zhao2021learning}         & 72.7\%            & 0.073           & 28.1   \\
        Xiong et al.~\cite{xiong2023towards}& \underline{73.8\%}            & \textbf{0.068}  & \underline{28.3}   \\ 
        GOPT (ours)                         & \textbf{76.1\%}   & \underline{0.070}           & \textbf{29.6}  \\ \midrule
        \textbf{\textit{Ablation studies}} \\
        GOPT w/o PG                         & 70.6\%            & 0.086           & 27.5   \\
        GOPT w/o IR                         & 73.2\%            & 0.078           & 28.5   \\
        GOPT w/o PT                         & 67.1\%            & 0.085           & 26.2   \\        
        GOPT-MLP                            & 67.8\%            & 0.079           & 26.4   \\        
        GOPT-GRU                            & 68.7\%            & 0.082           & 26.9   \\ \bottomrule       
    \end{tabular}
    \begin{tablenotes} [para,flushleft] 
        \footnotesize
        \item \textbf{Bold} indicates the best and \underline{underline} indicates the second best for that metric.
    \end{tablenotes}
    \end{threeparttable}
\end{table}

\begin{figure} [!t]  %
    \centering
    \includegraphics[width=\linewidth]{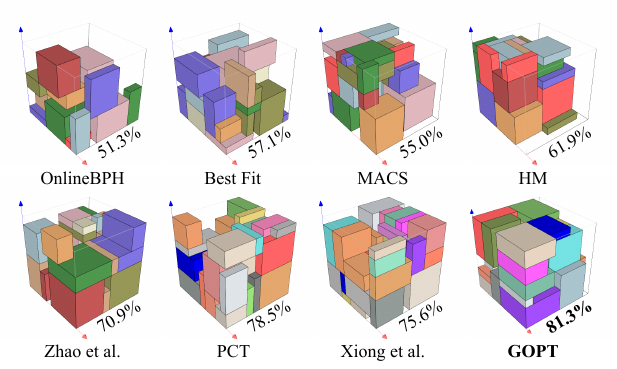}
    \vspace{-0.5cm}
    \caption{Visualization results of different methods for an item sequence in a $10\times10\times10$ bin. The number beside each bin indicates the value of \textit{Uti}.}
    \label{fig:visual}
    \vspace{-0.3cm}
\end{figure}

\begin{table*}[!ht]
    \centering
    \begin{threeparttable}
        \caption{Generalization performance on bins of different dimensions} 
        \label{tab:general_bin}
        \setlength{\tabcolsep}{5.5mm}
        \begin{tabular}{lcccccccc}
            \toprule
            \multirow{2}{*}{Method}              & \multicolumn{2}{c}{Bin-10}            & \multicolumn{2}{c}{Bin-30}            & \multicolumn{2}{c}{Bin-50}            & \multicolumn{2}{c}{Bin-100}  
            \\ \cmidrule(lr){2-3} \cmidrule(lr){4-5} \cmidrule(lr){6-7} \cmidrule(lr){8-9}
                                                 & \textit{Uti}       & \textit{Num}     &\textit{Uti}        & \textit{Num}     & \textit{Uti}       & \textit{Num}     & \textit{Uti}        & \textit{Num} 
            \\ \midrule
            Zhao et al. \cite{zhao2021online}    & 70.9\%             & 27.5             & 72.4\%             & 27.9             & 51.7\%             & 20.6             & /                   & /            \\
            Zhao et al. \cite{zhao2022learning}$^{1}$  & 70.1\%             & 27.1             & 71.7\%             & 27.7             & 72.6\%             & 28.1             & 71.3\%              & 27.6         \\
            PCT~\cite{zhao2021learning}          & 72.7\%             & 28.1             & 73.1\%             & 28.1             & 70.1\%             & 27.2             & 72.7\%              & 27.9         \\
            Xiong et al. \cite{xiong2023towards} & \underline{73.8\%} & \underline{28.3} & \underline{75.6\%} & 28.9             & 75.3\%             & 28.8             & 73.8\%              & 28.2         \\
            GOPT                                 & \textbf{76.1\%}    & \textbf{29.6}    & \textbf{76.0\%}    & \textbf{29.5}    & \underline{75.7\%} & \textbf{29.4}    & \underline{75.7\%}  & \underline{29.4} \\
            GOPT (Bin-10)$^{2}$                  & \textbf{76.1\%}    & \textbf{29.6}    & \textbf{76.0\%}    & \underline{29.2} & \textbf{75.8\%}    & \underline{29.2} & \textbf{76.3\%}     & \textbf{29.6}    \\ \bottomrule
        \end{tabular}
        \begin{tablenotes} 
            \item $^{1}$Results are copied directly from~\cite{zhao2022learning} since the code is not available.
            \item $^{2}$GOPT (Bin-10) refers to the GOPT policy trained in Bin-10, which we directly apply to four environments to obtain testing results. In contrast, the other four methods, along with GOPT, require separate training and testing in these environments.
        \end{tablenotes}
    \end{threeparttable}
    \vspace{-0.3cm}
\end{table*}

\subsubsection{Results}
We evaluate the packing performance of these methods using three metrics: average space utilization of the bin (\textit{Uti}), average number of packed items (\textit{Num}), and standard deviation of space utilization (\textit{Sta}), the latter of which assesses the stability of the methods across all instances. Quantitative comparisons, presented in \tabref{tab:result}, demonstrate that our method outperforms all baselines in terms of \textit{Uti} and \textit{Num}. The findings indicate that our method achieves superior item packing and more efficient utilization of bin space. It is noteworthy that our method achieves the second-highest performance in terms of \textit{Sta}, with DRL-based methods showing comparable performance in this metric. Moreover, all DRL-based methods significantly outshine heuristic methods across all evaluation metrics. This advantage is attributed to the DRL-based method's ability to extract packing patterns and regularities from extensive training samples. In contrast, heuristic methods may struggle to generalize beyond their specific rules or strategies. The comparison with the baselines indicates our method's effectiveness. Furthermore, we depict the qualitative comparisons of visualized packing results from different methods in \figref{fig:visual}. It is observed that our results are visually superior to other competing methods.

\subsection{Generalization }

The capacity of learning-based methods to generalize across diverse datasets and unseen scenarios has consistently been a subject of scrutiny and interest. 
This section evaluates the generalization performance of our method across various bins of different dimensions and unseen items.  

\textbf{Generalization on different bins.} In addition to the initial bin dimensions for the aforementioned training, we introduce three other environments where the bin dimensions are set to $30 \times 30 \times 30$, $50 \times 50 \times 50$, and $100 \times 100 \times 100$, respectively, and the item dimensions in the dataset are scaled up correspondingly. These environments are named Bin-10, Bin-30, Bin-50, and Bin-100. 
The search space for actions increases as the dimensions of bins grow, resulting in a higher complexity for finding a solution.
To assess our method's generalization ability regarding the bin dimensions, we directly transfer our policy, trained solely in Bin-10, to the other three environments without fine-tuning. We additionally train and test our proposed GOPT, along with several DRL-based baseline methods \cite{zhao2021online, zhao2022learning, zhao2021learning, xiong2023towards}, separately in different environments for greater persuasiveness. The results in terms of \textit{Uti} and \textit{Num} are summarized in \tabref{tab:general_bin}. It is noted that Zhao et al.'s method \cite{zhao2021online} fails to converge in Bin-100. According to \tabref{tab:general_bin}, GOPT not only maintains consistent performance across different environments but also consistently outperforms other methods. Significantly, the policy GOPT (Bin-10) without retraining shows stable performance in environments divergent from the training one. Other DRL-based methods do not possess such ability as they need to be retrained when encountering varying bin dimensions. 
Intriguingly, some of them achieve relatively high performance in Bin-30. We surmise that this is due to a balance between the increased number of model parameters and the moderate problem complexity for this size, allowing for enhanced fitting capacity without the excessive difficulty observed at larger bins.

\textbf{Generalization on unseen items.} Additionally, we conduct experiments to assess the generalization performance of our method using unseen items in Bin-10. This test is crucial and challenging as models often exhibit diminished performance when confronted with testing data that possess different characteristics. As previously mentioned, there are 125 distinct types of items in the RS dataset. We randomly exclude 25 types of items ($\text{RS}_{exc}$) from RS to train an agent with the sub-dataset $\text{RS}_{sub}$ and test it with the complete RS and $\text{RS}_{exc}$. We select two baselines that performed well in previous experiments for comparison. 
As shown in \tabref{tab:general_item}, our policy trained in the sub-dataset performs better than others when tested on both the full dataset RS and the dataset $\text{RS}_{exc}$ consisting entirely of unseen items. This suggests the trained policy exhibits adequate generalization ability even on unseen items.
We also observe an increase in \textit{Num} across all methods on RS and $\text{RS}_{exc}$, likely due to these datasets having more small, easier-to-pack items.

\begin{table}[!t]
    \centering
    \caption{Performance of policies trained on $\text{RS}_{sub}$ when evaluated on $\text{RS}_{sub}$ and two datasets containing unseen items}
    \label{tab:general_item}
    \begin{tabular}{lcccccc}
        \toprule
        \multirow{2}{*}{Method}              & \multicolumn{2}{c}{$\text{RS}_{sub}$}         & \multicolumn{2}{c}{RS}         & \multicolumn{2}{c}{$\text{RS}_{exc}$} \\ \cmidrule(lr){2-3} \cmidrule(lr){4-5} \cmidrule(lr){6-7}
                                             & \textit{Uti}    & \textit{Num} & \textit{Uti}    & \textit{Num} & \textit{Uti}    & \textit{Num}  \\ \midrule
        PCT~\cite{zhao2021learning}          & 73.9\%          & 28.0         & 73.7\%          & 28.2         & 73.7\%          & 29.3          \\
        Xiong et al.~\cite{xiong2023towards} & 73.8\%          & 27.9         & 73.0\%          & 27.8         & 72.9\%          & 29.0          \\ 
        \textbf{GOPT}                        & \textbf{75.5\%} & \textbf{28.7}& \textbf{76.1\%} & \textbf{29.5}& \textbf{75.7\%} & \textbf{30.2}  \\ \bottomrule 
    \end{tabular}
    \vspace{-0.3cm}
\end{table}

\subsection{Ablation Studies}

\begin{figure} [!t]
    \centering
    \includegraphics[width=0.8\linewidth]{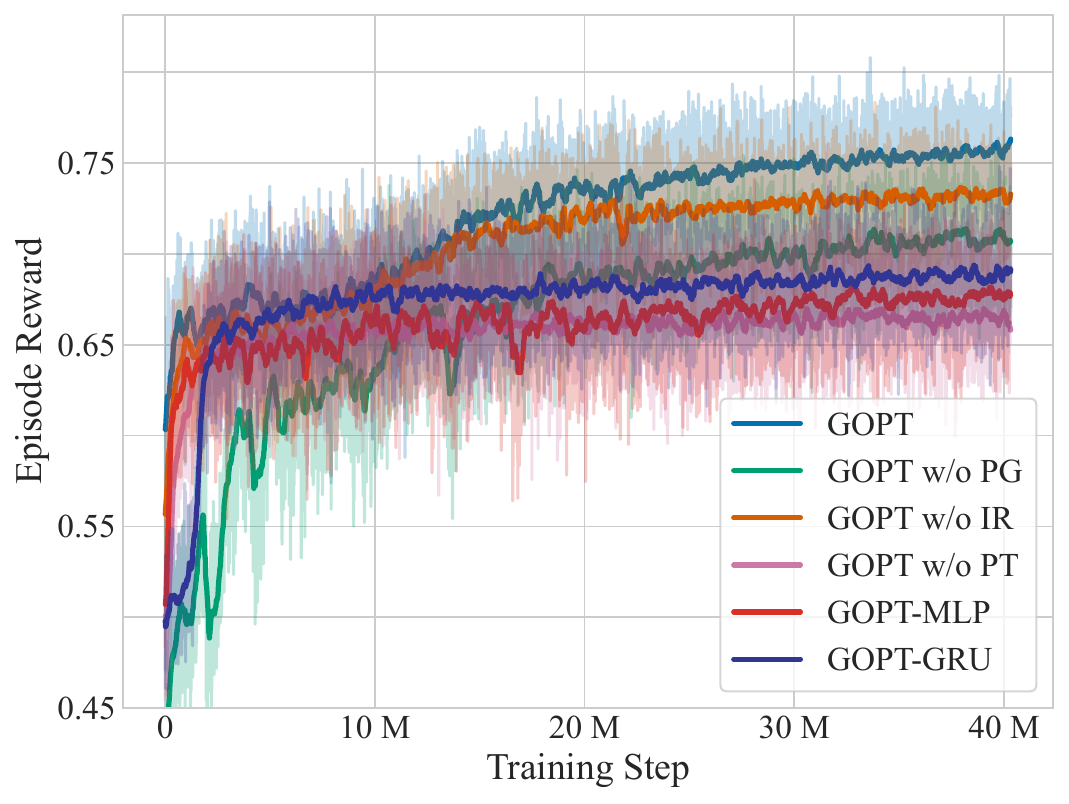}
    \vspace{-0.25cm}
    \caption{Comparison of the training performance for the ablation studies. The results are obtained with 128 different random seeds.}
    \label{fig:ablation}
    \vspace{-0.2cm}
\end{figure}

Additional ablation studies are conducted to thoroughly analyze the impact of various components in our method. These components encompass the Placement Generator (PG), item representation (IR), and Packing Transformer (PT). We exclude PG and provide the neural network with all the placements and the corresponding masks to elucidate its effect. We also present results obtained without transforming the item representation from a three-dimensional vector to the proposed mode. Additionally, we conduct experiments by removing PT (GOPT w/o PT) and replacing PT with MLP (GOPT-MLP) and GRU (GOPT-GRU) to gain insights into its significance. The results are depicted in \tabref{tab:result}. We also present reward curves versus training steps in \figref{fig:ablation}.

As shown in \tabref{tab:result} and \figref{fig:ablation}, all three components introduced in this study exhibit favorable outcomes in line with our expectations. The comparative analyses indicate the performance of GOPT w/o PT, GOPT-MLP, and GOPT-GRU is significantly degraded compared to GOPT. It highlights the advantageous role of identifying spatial relations through the proposed PT in enhancing performance. 
This capability can be attributed to the superior efficacy of the attention mechanism in handling intricate sequential data and in learning long-range dependencies compared to other networks.
Additionally, from \figref{fig:ablation}, the models incorporating PT (GOPT, GOPT w/o PG, GOPT w/o IR) require more training data to achieve convergence than the models without PT (GOPT w/o PT, GOPT-MLP, GOPT-GRU), approximately 30 million versus 10 million.
Besides, GOPT achieves greater space utilization and packs more items than GOPT w/o IR, indicating that the proposed item representation facilitates the DRL agent's learning and final performance. From \figref{fig:ablation}, we note that GOPT w/o PG attains the least reward during the initial stages of training. This suggests that the PG module informed by human experience can contribute to improving sampling efficiency when the DRL agent has yet to accumulate substantial packing knowledge. 

\begin{table}[!t]
    \centering
    \caption{Comparison of different reward functions}
    \label{tab:reward}
    \setlength{\tabcolsep}{5mm}
    \begin{tabular}{lccc}
        \toprule
         Reward designs                      &\textit{Uti}      &\textit{Num}  \\ \midrule
         Step-wise                           &\textbf{76.1\%}   &\textbf{29.6}  \\
         Terminal~\cite{xu2023neural}        &70.9\%            &27.6           \\ 
         Heuristic~\cite{yang2023heuristics} &72.4\%            &28.0           \\ \bottomrule
    \end{tabular}
    \vspace{-0.3cm}
\end{table}

We also investigate the impact of reward design for the problem, encompassing the step-wise reward employed in this work, the terminal reward~\cite{xu2023neural} defined as the final space utilization in an episode, and the heuristic reward~\cite{yang2023heuristics} which adds a penalty term to avoid wasted space due to unreasonable actions. According to \tabref{tab:reward}, the agent trained with the terminal reward shows the poorest performance, while the step-wise reward is more efficient despite its simpler and more intuitive nature than the heuristic reward.

\subsection{Real World Experiment}

\begin{figure} [!t]
    \centering
    \includegraphics[width=0.99\linewidth]{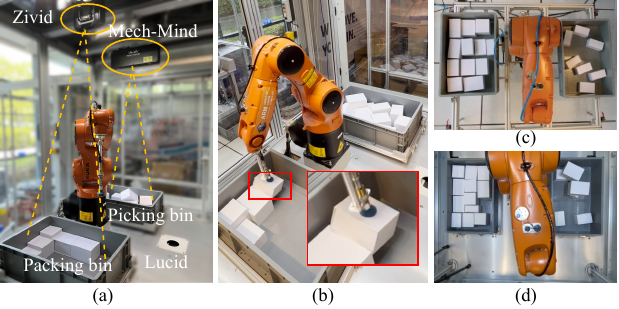}
    \caption{The real-world experiments.
        (a) Our robot packing setup: A KUKA robot is equipped with a suction cup and three 3D cameras;
        (b) Failure case: The primary sources of failure in our experiments are measurement errors;
        (c) and (d) are snapshots of safe packing and tight packing.}
    \label{fig:robot}
    \vspace{-0.3cm}
\end{figure}

We establish a physical robot packing testbed to verify the applicability of our method in the real world, as depicted in \figref{fig:robot}a. The dimensions of the bin for packing items are $56cm \times 36.5cm \times 21cm$, which is discretized into a bin of $80 \times 52 \times 30$, with each cell measuring $0.7cm$ in length. In this task, a robot selects a box from a bin, moves it within the Lucid camera's field of view to assess the box's dimensions and in-hand pose, and subsequently places it into another bin according to GOPT trained in the simulation. Meanwhile, two cameras are mounted to monitor these bins separately. The heightmap of the packing bin is generated through the segmentation and projection of the point cloud and the detection of rectangles. The pick-and-pack process proceeds until no boxes remain for picking or there is not enough space for packing the next box. Experiments show that a robot can utilize our method to complete the packing task in a real-world scenario. The demonstration video is provided in our supplementary materials.

From experiments, we observe that camera-induced measurement errors have the potential to cause collisions between boxes during placement (see \figref{fig:robot}b). To prevent this, an additional $0.7cm$ buffer space is allocated around each placed box, as shown in \figref{fig:robot}c, resulting in an average space utilization of 67.5\% across 20 tests. Reducing the buffer to zero increases the risk of errors and leads to 2 out of 20 tests failing, but achieves higher utilization ($73.3\%$ across 18 successful tests), as shown in \figref{fig:robot}d. 
These findings provide an impetus for future research aimed at enhancing both system reliability in real-world robotic packing scenarios and the compactness of the packing outcomes.

\section{Conclusions}

We contribute a novel framework called GOPT for online 3D bin packing. GOPT embraces the Placement Generator module to generate placement candidates and represent the state of a bin with these candidates. Meanwhile, the Packing Transformer identifies spatial correlations for packing, which employs a cross-attention mechanism to fuse information from items and the bin effectively. Extensive experiments prove GOPT's superiority over existing methods, demonstrating notable enhancements not only in packing performance but also in generalization capabilities. Specifically, trained GOPT policy can generalize both across varying bins and unseen items. 
Finally, we successfully apply the trained packing policy in a robotic system, demonstrating its practical applicability.
In the future, we plan to extend our method's application to include packing objects with irregular shapes, a common challenge in robotic pick-and-place tasks. We also plan to explore how to improve the reliability of the physical robot packing system.






\bibliographystyle{IEEEtran}
\bibliography{reference}

\end{document}